\documentclass[11pt]{article}

\usepackage[final]{acl}

\usepackage{times}
\usepackage{latexsym}

\usepackage[T1]{fontenc}

\usepackage[utf8]{inputenc}

\usepackage{microtype}

\usepackage{inconsolata}

\usepackage{graphicx}
\usepackage{booktabs}
\usepackage{makecell}
\usepackage{amsmath}
\usepackage{listings}
\usepackage{tcolorbox}
\usepackage{xcolor}
\usepackage{fontenc}
\usepackage{lmodern}
\usepackage{bbm}
\usepackage{todonotes}
\usepackage{multirow}
\usepackage{makecell}
\tcbuselibrary{listings, skins, breakable}
\DeclareMathOperator*{\E}{\mathbbm{E}}
\title{Notes to Self: Can LLMs Benefit from Experiential Abstractions?}

\author{Chang Liu, Xinyu Li, \and Artur Dubrawski \\
  Auton Lab, Carnegie Mellon University \\
  \texttt{\{changl8,xinyul2,awd\}@cs.cmu.edu}}

\begin{document}
\maketitle
\begin{abstract}
Humans distill experience into reusable \textit{abstractions}, e.g., strategies and cautionary reminders, and apply them to gradually solve problems more effectively.
We study whether Large Language Models (LLMs) can similarly benefit from such experiential abstractions.
From LLMs' solution traces on a training set, a stronger teacher or the LLMs themselves extract natural-language abstractions into a retrievable library.
We explore two modes of abstraction usage: (1) inference-time retrieval and (2) reinforcement learning (RL) with abstraction-augmented training prompts.
Experiential abstractions improve LLM performance on mathematical and logical reasoning benchmarks.
Self-extracted abstractions match teacher-extracted ones, and our abstraction usage framework can transfer to other datasets and models.
These findings suggest LLMs can extract and apply experiential abstractions much as humans leverage distilled experience.
\end{abstract}

\section{Introduction}
Humans actively filter, compress, and internalize past experiences into reusable \textit{abstractions}, e.g., heuristics, strategies, and cautionary reminders~\citep{zsambok2014naturalistic}, rather than solving every new problem from scratch or merely accumulating memory: a student compiles a cheatsheet of recurring tricks; a technician carries a mental catalog of techniques.
These experiential abstractions act as cognitive toolkits for increasingly complex tasks~\citep{gick1983schema, polya1945solve}.
As large language models (LLMs) advance on complex reasoning tasks such as mathematics~\citep{shao2024deepseekmath}, we ask an analogous question: can they similarly benefit from experiential abstractions derived from their own reasoning experience?

Recent work explores experiential learning in LLMs~\citep{zhao2024expel, suzgun2026dynamic, didolkar2025metacognitive, qu2025rlad, xia2026skillrl}, but typically relies on large frontier models as the target or as a teacher supervising smaller students, leaving unclear whether student LLMs can themselves extract or use such abstractions.

\begin{figure*}[t]
  \centering
  \includegraphics[width=0.86\textwidth]{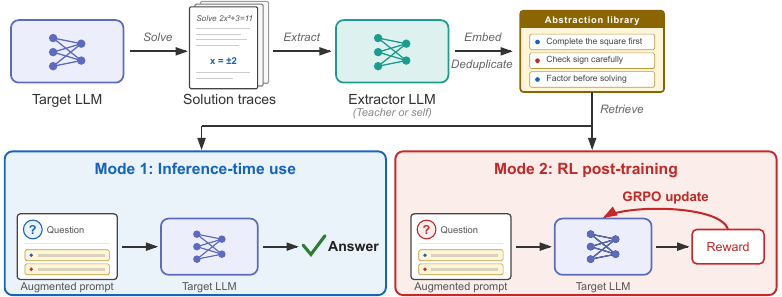}
  \caption{\textbf{Pipeline Overview.} An extractor LLM (frontier teacher or itself) distills the target LLM's training traces into a library of natural-language abstractions, used in two modes: (1) inference-time retrieval into the test prompt, and (2) GRPO~\citep{shao2024deepseekmath} post-training on abstraction-augmented prompts.}
  \label{fig:overview}
\end{figure*}

We assess the opportunity of bridging this gap by deriving a library of experiential abstractions from a target LLM's own training traces, with extraction by either a stronger teacher model or the target LLM itself.
We study two modes of abstraction use (Figure~\ref{fig:overview}): (1) inference-time retrieval into the test prompt, and (2) RL fine-tuning with abstraction-augmented training prompts.

We find that experiential abstractions extracted from the \texttt{MATH}~\citep{hendrycks2021measuring} training set can improve LLM mathematical reasoning abilities on \texttt{MATH-500}~\citep{lightman2024let} for both modes of abstraction usage, with consistent gains from abstraction-augmented RL post-training.
Moreover, self-extracted abstractions yield gains comparable to teacher-extracted ones.
Additionally, our abstraction-usage pipeline can transfer to other datasets and domains (e.g., logical reasoning), and abstractions from one model's traces can benefit a different target.
These results suggest that LLMs can effectively extract and apply experiential abstractions from their own reasoning experience, incrementally self-improving their abilities as their use expands. 
The code of this paper is available at 
\url{https://github.com/autonlab/Notes-to-self}.

\section{Related work}
\label{sec:related_work}
Several recent works focusing on eliciting LLMs' ability to learn from experience heavily rely on large or frontier LLMs.
ExpeL~\citep{zhao2024expel} extracts insights from agent trajectories and applies them at inference to a frozen frontier LLM in agentic environments. 
Dynamic Cheatsheet~\citep{suzgun2026dynamic} maintains a test-time-evolving memory for black-box frontier LLMs across reasoning queries. 
Metacognitive Reuse~\citep{didolkar2025metacognitive} extracts named behaviors from a 70B model's own traces and uses them in-context, in self-improvement loops, or as supervised fine-tuning on behavior-conditioned traces. 
RLAD~\citep{qu2025rlad} co-trains an abstraction generator alongside the solver via two-player RL, producing abstractions on the fly per problem, yet uses a strong teacher model to supervise the abstraction generator. 
SkillRL~\citep{xia2026skillrl} maintains a hierarchical skill library compiled by a frontier teacher model, applies cold-start SFT to teach abstraction use, then runs RL with skills evolving during training, in agentic settings. 

In this work, we ask whether small student LLMs themselves can extract and benefit from such abstractions through outcome-reward RL alone, without a frontier model at inference or teacher demonstrations for training.
Moreover, our approach differs from reflection-style methods~\citep{shinn2023reflexion,madaan2023self}, which are a form of test-time scaling, leveraging multiple feedback and regeneration cycles to improve performance with additional test-time compute. Our approach instead emphasizes train-to-test knowledge transfer, distilling the insights from training into a library of abstractions reusable on unseen test instances.
\section{Methods}
\label{sec:methods}
We denote the target LLM by $\pi_\theta$ with parameters $\theta$, its solution trace on a question $q$ by $y \sim \pi_\theta(\cdot \mid q)$, and the abstraction library by $\mathcal{L} = \{a_1, \ldots, a_N\}$, where each $a_i$ is a natural-language abstraction. A sentence embedder $\phi(\cdot)$ maps both questions and abstractions to a shared vector space for retrieval. We use outcome reward $r(y, y^*) \in \{0, 1\}$, set to one when the final answer in $y$ matches the ground truth $y^*$.
Figure~\ref{fig:overview} summarizes the full pipeline: we build a library of experiential abstractions from a target LLM's training traces and study two modes of using it.

\subsection{Building the library of experiential abstractions}
\label{sec:methods-library}
We first collect solution traces of the target LLM on a training set (e.g., the \texttt{MATH}~\citep{hendrycks2021measuring} training split of 7{,}500 problems) and sample one trace per problem from $\pi_\theta$. Each trace is labeled correct or incorrect by comparing its final answer with the ground truth.

We then prompt an extractor, either the target LLM itself or a stronger teacher, to summarize each trace into one or two reusable abstractions. 
Given the problem, the solution trace, the ground-truth answer, and the correctness label, the extractor is instructed to produce single-sentence, reusable abstractions that are free of problem-specific numbers, variable names, or the answer itself. 

Extraction is conditioned on correctness: for correctly solved problems the extractor identifies key reasoning moves worth reusing, producing \emph{strategy} abstractions, while for failed problems it identifies mistake patterns to avoid, producing \emph{caution} abstractions. 
We show the full extraction prompts in Appendix~\ref{app:prompt-extract-abs}.

To construct the library, we embed each abstraction with the \texttt{all-MiniLM-L6-v2} sentence encoder~\citep{reimers2019sentence} into 384-dimensional $\ell_2$-normalized vectors. We remove near-duplicates by clustering abstractions with cosine-similarity union-find at a threshold of $0.85$, keeping strategy and caution entries in separate buckets, and storing the surviving entries.
This deduplication scheme effectively reduces the library size by approximately 30\%, as shown in Table~\ref{tab:dedup} in Appendix~\ref{app:dedup}. 

\subsection{Inference-time use of abstractions}
\label{sec:inference}
To use abstractions for inference, we retrieve relevant abstractions from the library for each test question and inject them into the prompt. Given the query representation $\tilde{q}$ of a question $q$, we retrieve top-k (k=6 in our experiments) similar library entries by cosine similarity:
\begin{equation}
\mathcal{R}_q = \operatorname*{top\text{-}k}_{a \in \mathcal{L}}\ \cos\bigl(\phi(\tilde{q}), \phi(a)\bigr).
\end{equation}
We construct $\tilde{q}$ in one of two ways. The default query prepends the problem's subject tag to its text, ``[\{subject\}] \{question\}'', and falls back to the plain question when no subject is available. We additionally consider a rewritten query, obtained by prompting the LLM to restate the problem as a single abstraction-style sentence (Appendix~\ref{app:prompt-rewrite-input}). Retrieval is score-gated: we retrieve with both queries and adopt the rewritten one only when its top-1 cosine score exceeds the original's by more than $0.02$, breaking ties in favor of the original.

The retrieved set $\mathcal{R}_q$ is grouped by type (strategy/caution) and inserted into the prompt as a \texttt{<reference\_notes>} block, placed after the question. The LLM then generates its solution from the augmented prompt $x = \mathrm{Prompt}(q, \mathcal{R}_q)$ in a single forward pass (see Appendix~\ref{app:prompt-abs-rollout} for the prompts).

\subsection{Abstraction-augmented RL post-training}
Here, abstractions enter during RL fine-tuning rather than only at inference. We apply the above retrieval procedure to each training question $q$, obtaining its abstraction set $\mathcal{R}_q$, and fine-tune the target LLM with GRPO~\citep{shao2024deepseekmath} so that the policy conditions on the pair $(q, \mathcal{R}_q)$ rather than on $q$ alone.

For each training question, GRPO samples a group of candidate solutions $y \sim \pi_\theta(\cdot \mid q, \mathcal{R}_q)$, scores each with the outcome reward $r(y, y^*)$, and normalizes rewards within the group to form the advantage $\hat{A}(q, y)$. The policy is updated to maximize
\begin{equation}
\mathcal{J}(\theta) = \E\limits_{\substack{q \sim \mathcal{D} \\ y \sim \pi_\theta(\cdot \mid q, \mathcal{R}_q)}}\bigl[\hat{A}(q, y)\, \log \pi_\theta(y \mid q, \mathcal{R}_q)\bigr],
\end{equation}
where $\mathcal{D}$ is the set of training questions and $\pi_\theta(\cdot \mid q, \mathcal{R}_q)$ denotes the policy applied to the prompt that combines $q$ with its retrieved abstractions (Section~\ref{sec:inference}). 
In short, the reward signal and optimization match those of standard GRPO, while the policy is now conditioned on $(q, \mathcal{R}_q)$. 
After training, the model can be evaluated with or without abstraction-augmented prompts at test time.
\section{Results}
\label{sec:results}

We evaluate two open-source LLMs \texttt{Llama-3.2-3B-Instruct}~\citep{grattafiori2024llama} and \texttt{Qwen-2.5-1.5B-Instruct}~\citep{qwen2025qwen25technicalreport}, whose quantitative reasoning abilities are less saturated than newer models.
We adopt \texttt{DeepSeek-V4-Flash}~\citep{deepseekv4} as the teacher model for abstraction extraction.
We report \texttt{pass@1/pass@8}, averaged over $8$ rollouts at temperature $0.6$. 


\begin{table*}[ht]
\centering
\setlength{\tabcolsep}{3pt}
\caption{Performance (\texttt{pass@1/pass@8}) on \texttt{MATH-500} across five configurations of abstraction usage, and teacher- vs. self- extracted abstractions.}
\label{tab:main-results}
{\footnotesize
\begin{tabular}{@{}lcccccc@{}}
\toprule
Model & \makecell[l]{Abstraction extractor} &
  Baseline & Inference$_{\text{abs}}$ & GRPO & GRPO$_{\text{train}}$ & GRPO$_{\text{train+test}}$ \\ \midrule
\multirow{2}{*}{\texttt{Llama}}
  & Teacher & \multirow{2}{*}{43.11/70.57} & 46.42/\textbf{72.80} & \multirow{2}{*}{46.89/70.54} & \textbf{49.45}/71.98 & 47.65/70.23 \\
  & Self    &                             & 46.30/71.66 &                             & 49.07/71.14 & 47.67/70.69 \\
\midrule
\multirow{2}{*}{\texttt{Qwen}}
  & Teacher & \multirow{2}{*}{50.23/76.47} & 49.14/75.09 & \multirow{2}{*}{52.55/75.55} & 53.44/76.34 & 52.66/75.76 \\
  & Self    &                             & 49.44/74.92 &                             & \textbf{53.77}/\textbf{76.83} & 52.72/75.22 \\
\bottomrule
\end{tabular}%
}
\end{table*}

\begin{table}[ht]
\centering
\setlength{\tabcolsep}{3pt}
\caption{Ablation isolating prompt-template effects on \texttt{MATH-500} (\texttt{pass@1/pass@8}), using \texttt{Llama-3.2-3B-Instruct} with teacher-extracted abstractions. \emph{No notes} is the plain configuration; \emph{Blank notes} inherits the same prompt structure but leaves the \texttt{<reference\_notes>} block empty; + \emph{Abstractions} fills it with retrieved abstractions.}
\label{tab:ablation}
\resizebox{\linewidth}{!}{%
\begin{tabular}{@{}lccc@{}}
\toprule
Mode & No notes & Blank notes & + Abstractions \\ \midrule
Inference-time &
  \makecell{Baseline\\43.11/70.57} &
  \makecell{Inference$_{\text{blank}}$\\46.02/70.00} &
  \makecell{Inference$_{\text{abs}}$\\{46.42/72.80}} \\ \midrule
RL (GRPO) &
  \makecell{GRPO\\46.89/70.54} &
  \makecell{GRPO$_{\text{blank}}$\\48.55/70.98} &
  \makecell{GRPO$_{\text{train}}$\\{49.45/71.98}} \\
\bottomrule
\end{tabular}%
}
\end{table}

\paragraph{Experiential abstractions improve mathematical reasoning.}
We first evaluate on the \texttt{MATH-500} test set using \texttt{DeepSeek-V4-Flash}-extracted abstractions from the MATH training split and compare five configurations of abstraction usage: Baseline (no fine-tuning or abstractions); Inference$_\text{abs}$ (abstractions injected at test time only); GRPO (vanilla GRPO, no abstractions); GRPO$_\text{train}$ (GRPO with abstractions injected in training prompts only); and GRPO$_\text{train+test}$ (abstractions injected in both training and test prompts).

As shown in Table~\ref{tab:main-results}, inference-time injection alone can be helpful: it raises \texttt{Llama}'s \texttt{pass@1} by $+3.31$ and \texttt{pass@8} by $+2.23$ over the baseline, though slightly lowering both for \texttt{Qwen}, likely due to answer formatting and the quality of retrieved abstractions (Appendix~\ref{app:qwen-analysis}).
Abstraction-augmented RL yields more consistent gains: GRPO$_{\text{train}}$ attains $+2.56$ over vanilla GRPO and $+6.34$ over baseline for \texttt{Llama} and $+0.89$ and $+3.21$ for \texttt{Qwen}, respectively.
Notably, across all usage modes and models, self-extracted abstractions perform comparably with teacher-extracted ones, indicating that a small LLM can distill useful experiential abstractions from its own traces without supervision from a frontier model.

Interestingly, GRPO$_{\text{train+test}}$ underperforms GRPO$_{\text{train}}$, which we attribute to RL sharpening the model output distribution towards near-deterministic correct modes~\citep{yue2025doesreinforcementlearningreally}, on which the perturbation effect of abstractions can hurt performance (Appendix~\ref{app:effect-of-rl}).

We further perform an ablation study where models inherit the prompt structure of training/inference time injection but leave the \texttt{<reference\_notes>} abstractions empty (Table~\ref{tab:ablation}).
Both GRPO$_{\text{train}}$ and Inference$_{\text{abs}}$ outperform their blank controls on pass@1 and pass@8, showing the abstractions' unique advantage in extending model capabilities.

\paragraph{Abstraction-usage pipeline transfers to other datasets and models.}
\begin{table*}[ht]
\centering
\setlength{\tabcolsep}{3pt}
\caption{Out-of-distribution results (\texttt{pass@1/pass@8}) with \texttt{Llama-3.2-3B-Instruct}, spanning easier and harder math (\texttt{GSM8K}, \texttt{OlympiadBench}) and logical reasoning (\texttt{MuSR}). For \texttt{MuSR}, we report checkpoints trained (and abstractions extracted) on \texttt{MATH} or \texttt{FOLIO}.}
\label{tab:ood}
{%
\footnotesize
\begin{tabular}{@{}lcccc@{}}
\toprule
Test set & Baseline & GRPO & GRPO$_{\text{blank}}$ & GRPO$_{\text{train}}$ \\ \midrule
\texttt{GSM8K}                 & 54.82/85.44 & 77.34/92.49 & \textbf{81.30}/\textbf{93.33} & 80.48/92.95 \\
\texttt{\makecell[l]{OlympiadBench}} & 13.91/34.57 & 16.04/36.94 & 16.36/35.91 & \textbf{17.34}/\textbf{37.24} \\ \midrule
\texttt{MuSR\textsubscript{MATH}}  & \multirow{2}{*}{40.94/78.70} & \textbf{47.06}/75.79 & 45.21/76.59 & 46.20/75.93 \\
\texttt{MuSR\textsubscript{FOLIO}} &                              & 44.58/75.00 & 44.82/77.91 & 45.54/\textbf{79.37} \\
\bottomrule
\end{tabular}%
}
\end{table*}

\begin{table}[ht]
\centering
\setlength{\tabcolsep}{3pt}
\caption{Cross-model transfer of abstractions extracted from \texttt{Llama-3.2-3B-Instruct} traces (teacher-extracted), injected at inference into other model families on \texttt{MATH-500} (\texttt{pass@1/pass@8}).}
\label{tab:transfer}
\resizebox{\linewidth}{!}{%
\begin{tabular}{@{}lccc@{}}
\toprule
Target model & Baseline & Inference$_{\text{blank}}$ & Inference$_{\text{abs}}$ \\ \midrule
\texttt{gemma-2-2B-it}          & 21.77/43.00 & 13.08/33.60 & 13.75/33.40 \\
\texttt{gemma-3-3B-it}          & 44.88/62.40 & 42.02/62.40 & 40.15/60.00 \\
\texttt{\makecell[l]{Phi-3.5-mini\\-instruct}}  & 40.62/64.00 & 42.30/65.20 & 43.60/65.80 \\
\bottomrule
\end{tabular}%
}
\end{table}
We further assess how our abstraction-usage pipeline transfers to other datasets, domains, and models. 
We use the \texttt{DeepSeek} teacher for abstraction extraction in these experiments. 

We first find that MATH-derived abstractions and model checkpoints induce gains on the harder \texttt{OlympiadBench}~\citep{he2024olympiadbench} but not the easier \texttt{GSM8K}~\citep{cobbe2021training} (Table~\ref{tab:ood}), suggesting that abstractions help more on problems the model is less capable.
This is also corroborated by further analyses on the \texttt{MATH-500} results (Appendix~\ref{app:effect-of-rl}). 

On the logical reasoning benchmark \texttt{MuSR}~\citep{sprague2024musr}, we observe that while MATH can only provide useful RLVR signals (pass@1:  GRPO > GRPO$_{\text{train}}$ > Baseline), a logical reasoning dataset \texttt{FOLIO}~\citep{han2024folio} can provide informative abstractions and extend model capability by inducing gains in pass@8 (79.37, GRPO$_{\text{train}}$, best configuration).

Lastly, \texttt{Llama}-derived abstractions can be transferred to \texttt{Phi-3.5-mini-instruct}~\citep{abdin2024phi3technicalreporthighly}, but the blank template itself already harms \texttt{Gemma} models~\citep{team2024gemma,gemmateam2025gemma3technicalreport} (Table~\ref{tab:transfer}), suggesting that transferability of abstractions is specific to the downstream model.
\section{Conclusion}
\label{sec:conclusion}
We studied whether LLMs benefit from experiential abstractions distilled from training traces, via either inference-time retrieval or abstraction-augmented RL post-training.
Such abstractions extend model capability across mathematical and logical reasoning benchmarks.
Moreover, small student LLMs extract abstractions with comparable effectiveness to frontier teachers, and abstractions can transfer across these target models.
These findings indicate that experiential abstraction is a viable self-improvement signal for LLMs, mirroring how humans distill and reuse past experience.

\section*{Limitations}

Our results depend on several fixed design choices: a particular teacher model, sentence encoder, retrieval cutoff $k$, and deduplication threshold, for which we do not provide a sensitivity analysis. Broader cross-model transfer claims would require additional pairs. RL post-training is conducted under a modest compute budget (one epoch, eight rollouts per prompt), and scaling the training could change the magnitude or ordering of the observed gains.

Our method targets mathematical problem solving and extends to logical reasoning, and we do not foresee direct societal risks. The main caveat is that an abstraction library can encode incorrect or misleading heuristics; outcome-reward RL mitigates this by reinforcing only verified-correct solutions.

\section*{Acknowledgments}

This work was partially supported by the National Science Foundation awards 2406231 and 2530752, and by the US Department of Energy award DE-SC0026594.

\bibliography{custom}

\clearpage
\appendix
\section{Interpreting the effect of RL and abstractions}
\label{app:effect-of-rl}

In the main text (Section~\ref{sec:results}) we observe that GRPO$_{\text{train+test}}$ consistently underperforms GRPO$_{\text{train}}$: re-injecting abstractions at test time hurts a model already fine-tuned with them. 
We attribute this to the interaction between abstractions and RL, and here we test two competing explanations. 
Under an \emph{internalization} hypothesis, the fine-tuned model has already absorbed the abstractions and is merely distracted by their re-injection. 
Under an \emph{exploration-vs-sharpening} account, abstractions perturb the output distribution to aid exploration, while RL sharpens that distribution toward near-deterministic correct modes, so perturbing an already-tuned model displaces probability mass away from correct solutions.

\paragraph{The internalization hypothesis does not hold.}
We evaluate three checkpoints---the base model (Baseline), plain GRPO (trained without abstractions), and GRPO$_{\text{train}}$---each with and without abstractions injected at test time (Table~\ref{tab:internalization}). 
Internalization predicts that test-time injection should harm only GRPO$_{\text{train}}$, the sole checkpoint exposed to abstractions during training. 
Instead, injection also degrades plain GRPO, which never encountered abstractions in training and therefore cannot have
internalized them. Internalization alone thus cannot explain the drop.

\begin{table}[ht]
\centering
\setlength{\tabcolsep}{6pt}
\caption{Effect of injecting abstractions at test time (\texttt{pass@1}) on \texttt{Llama-3.2-3B-Instruct} across three checkpoints. Injection degrades both RL-tuned checkpoints, including plain GRPO, which never saw abstractions during training.}
\label{tab:internalization}
\resizebox{\linewidth}{!}{
\begin{tabular}{@{}lccc@{}}
\toprule
Checkpoint & w/o abs.\ & w/ abs.\ & $\Delta$ \\ \midrule
Baseline              & 43.11 & 46.42 & $+3.31$ \\
GRPO                  & 46.89 & 45.76 & $-1.13$ \\
GRPO$_{\text{train}}$ & 49.45 & 47.65 & $-1.80$ \\
\bottomrule
\end{tabular}
}
\end{table}

\paragraph{Abstractions help where the model is weak and hurt where it is strong.}
To locate the effect, we bucket each checkpoint's 500 \texttt{MATH-500} problems by their per-problem accuracy \emph{before} injection, then measure the mean change in \texttt{pass@1} \emph{after} inference-time injection within each bucket (Table~\ref{tab:acc-buckets}).
In the low-accuracy bucket ($0$--$50\%$), where all three checkpoints struggle, abstractions help every checkpoint. 
In the mid bucket ($50$--$100\%$), injection still helps the base model but hurts both RL-tuned checkpoints; in the fully-solved bucket ($100\%$) it hurts all three.
Abstractions thus aid the model on problems it is less capable of---encouraging exploration---but are detrimental once it is already competent. 
This directly corroborates the main-text observation that MATH-derived abstractions help the harder OlympiadBench (where the model is weak) but not the saturated GSM8K (where it is strong), in Table~\ref{tab:ood}.

\begin{table}[ht]
\centering
\setlength{\tabcolsep}{6pt}
\caption{Mean change in \texttt{pass@1} after inference-time abstraction injection, with problems bucketed by pre-injection per-problem accuracy. Abstractions help low-accuracy problems but hurt high-accuracy ones, increasingly so for RL-tuned checkpoints.}
\label{tab:acc-buckets}
\resizebox{\linewidth}{!}{
\begin{tabular}{@{}lccc@{}}
\toprule
Accuracy bucket & Baseline & GRPO & GRPO$_{\text{train}}$ \\ \midrule
$0$--$50\%$   & $+6.31$ & $+3.62$ & $+0.83$ \\
$50$--$100\%$ & $+2.46$ & $-5.22$ & $-3.61$ \\
$100\%$       & $-2.32$ & $-5.29$ & $-1.12$ \\
\bottomrule
\end{tabular}
}
\end{table}

\paragraph{Exploration versus sharpening.}
These patterns are consistent with abstractions acting as a perturbation of the model's output distribution. 
For the untrained base model, whose distribution is comparatively diffuse, this perturbation broadens the search and raises coverage: injecting abstractions at test time lifts the base model's \texttt{pass@8} by $+2.23$ (Table~\ref{tab:ablation}). 
RLVR, however, is known to sharpen the output distribution toward near-deterministic correct modes \citep{yue2025doesreinforcementlearningreally}. 
Once the model is RL-tuned, these modes already concentrate on the correct answers for problems it can solve, so perturbing them via test-time injection can only displace mass away from correct solutions---explaining why injection hurts RL-tuned checkpoints most on problems they already solve, and why \texttt{pass@8} gains from abstraction content shrink after RL tuning (e.g., \textbf{GRPO}$_{\text{train}}$ raises \texttt{pass@8} by $+1.44$ over vanilla \textbf{GRPO}, versus the base model's $+2.23$). Our logical-reasoning results reinforce this: on \texttt{MuSR} (Table~\ref{tab:ood}), \textbf{GRPO}$_{\text{train}}$ is the only configuration that recovers the \texttt{pass@8} drop incurred by both \textbf{GRPO} and \textbf{GRPO}$_{\text{blank}}$.
\section{Quantitative analyses on \texttt{Qwen-2.5-1.5B-Instruct} results}
\label{app:qwen-analysis}

Inference-time abstraction injection improves \texttt{Llama-3.2-3B-Instruct} but slightly degrades \texttt{Qwen-2.5-1.5B-Instruct} ($50.23 \to 49.14$ \texttt{pass@1};
Table~\ref{tab:main-results}). 
We analyze the trajectories of Qwen's Inference$_{\text{abs}}$ against its Baseline (standard chain-of-thought prompt) across all 500 \texttt{MATH-500} problems, and find that degradation is driven by \emph{format disruption and poor retrieval fit} rather than by retrieved cautions acting as negative constraints.

\paragraph{Retrieved cautions do not act as negative constraints.}
A natural hypothesis is that retrieved \emph{caution} notes over-constrain \texttt{Qwen} and suppress its standard reasoning; if so, retrieving more cautions should worsen performance. 
We observe the opposite: the number of cautions retrieved is not positively correlated with degradation (Pearson $\rho = +0.069$, $p = 0.13$), and problems with more retrieved cautions are, if anything, less degraded (Table~\ref{tab:qwen-cautions}).

\begin{table}[ht]
\centering
\setlength{\tabcolsep}{6pt}
\caption{Mean change in \texttt{pass@1} (Inference$_{\text{abs}}$ $-$
Baseline) for \texttt{Qwen-2.5-1.5B-Instruct}, grouped by the number of caution notes retrieved. More cautions do not induce more degradation.}
\label{tab:qwen-cautions}
\resizebox{\linewidth}{!}{
\begin{tabular}{@{}ccc@{}}
\toprule
\# Cautions retrieved & \# Problems & Mean $\Delta(\texttt{pass@1})$ \\ \midrule
$0$--$1$ & 143 & $-2.08$ \\
$2$--$3$ & 233 & $-1.15$ \\
$4$--$6$ & 124 & $+0.15$ \\ \midrule
Total    & 500 & $-1.09$ \\
\bottomrule
\end{tabular}
}
\end{table}

\paragraph{Format sensitivity.}
Comparing per-problem trajectories, abstraction injection improves \texttt{Llama}'s answer-format compliance but harms \texttt{Qwen}'s: the rate of missing \texttt{\textbackslash boxed\{\}} answers falls for \texttt{Llama} ($7.3\% \to 4.2\%$) but rises for \texttt{Qwen} ($3.8\% \to 4.6\%$). 
This disruption alone accounts for \texttt{Qwen}'s most-degraded problems. For instance, on \texttt{Qwen}'s single most-degraded problem (\href{https://huggingface.co/datasets/HuggingFaceH4/MATH-500/viewer/default/test?row=122}{id~1332}), $24/32$ \textbf{Inference}$_{\text{abs}}$ samples derive the correct answer but omit the boxed format and are scored incorrect (Baseline: $11/32$ correct); no cautions are retrieved and the chain of thought remains intact, isolating the failure to formatting.

\paragraph{Retrieval fit.}
Degradation also arises when retrieval surfaces abstractions that fit the query poorly. 
On a problem asking for the least possible sum of distinct positive integers with a fixed product (\href{https://huggingface.co/datasets/HuggingFaceH4/MATH-500/viewer/default/test?row=309}{id~203}), \texttt{Llama} retrieves an apt minimum-sum strategy (``distribute the prime factors so that the resulting numbers are as close in magnitude...'') and adopts it near-verbatim, rising from $22\%$ to $63\%$. 
\texttt{Qwen} instead retrieves a strategy for the \emph{opposite} objective (``To maximize the sum of distinct positive integers with a fixed product...''), and its failing samples follow that prescription, dropping from $50\%$ to $16\%$. 

Together, these analyses indicate that \texttt{Qwen}'s degradation under inference-time injection stems from disrupted answer formatting and occasional poorly-matched retrievals, not from cautions functioning as harmful constraints.
\section{Effect of library deduplication}
\label{app:dedup}
Table~\ref{tab:dedup} shows the effect of deduplicating the abstraction library as described in Section~\ref{sec:methods-library}.
\begin{table}[ht]
\centering
\caption{The effect of library deduplication process. Each cell corresponds to the size of the library before$\to$after deduplication, for a target model-extractor pair. }
\label{tab:dedup}
\resizebox{\linewidth}{!}{%
\begin{tabular}{@{}ccc@{}}
\toprule
                                & Teacher         & Self           \\ \midrule
\texttt{Llama-3.2-3B-Instruct}  & $10915\to 7676$ & $6727\to 4232$ \\
\texttt{Qwen-2.5-1.5B-Instruct} & $11332\to 8274$ & $5647\to 3561$ \\ \bottomrule
\end{tabular}%
}
\end{table}
\section{Implementation details}
\subsection{Hardware compute}
All experiments are run on an NVIDIA DGX Station A100 machine, using all 4 NVIDIA A100 80GB GPUs on the DGX Station. 
Each single training run on the \texttt{MATH} dataset takes about 2.5 hours. 
Each inference run on the \texttt{MATH-500} dataset takes about 0.3 hours. 
Extracting \texttt{MATH} traces takes about 0.5 hours.

\subsection{Teacher model access}
We access the \texttt{DeepSeek-V4-Flash} teacher model via API granted by OpenRouter.

\subsection{Hyperparameters}
\label{app:hyperparams}
We report the training and inference hyperparameter configurations in Table~\ref{tab:train-hparams} and~\ref{tab:eval-hparams}. 
These configurations persist across all used models.

\subsection{Packages}
GRPO training is implemented with \texttt{verl}~\citep{sheng2024hybridflow}. 
Final-answer correctness against the ground truth is determined by extracting the ``\textbackslash boxed'' block and string-matching it against the ground truth.
We did not perform a hyperparameter search; the values in Tables~\ref{tab:train-hparams} and~\ref{tab:eval-hparams} follow standard GRPO settings.



\subsection{Use of AI assistants}
We used Claude (Anthropic) via the web interface for assistance with writing and editing, and the Cursor IDE (with an integrated LLM assistant) for coding support.

\begin{table}[h]
\centering
\caption{Training hyperparameters (GRPO).}
\label{tab:train-hparams}
\begin{tabular}{lc}
\toprule
\textbf{Hyperparameter} & \textbf{Value} \\
\midrule
\multicolumn{2}{l}{\textit{Model \& Data}} \\
Max prompt length    & 1536 \\
Max response length  & 3072 \\
Training batch size  & 128 \\
\midrule
\multicolumn{2}{l}{\textit{Optimization}} \\
Learning rate        & $1 \times 10^{-6}$ \\
LR schedule          & Cosine \\
Warmup ratio         & 0.1 \\
Epochs               & 1 \\
PPO mini-batch size  & 128 \\
\midrule
\multicolumn{2}{l}{\textit{Algorithm (GRPO)}} \\
Advantage estimator  & GRPO \\
KL loss coefficient  & $5 \times 10^{-4}$ \\
KL loss type         & Low-variance KL \\
KL in reward         & \texttimes \\
Entropy coefficient  & 0 \\
Rollouts per prompt  & 8 \\
\bottomrule
\end{tabular}
\end{table}
\begin{table}[h]
\centering
\caption{Evaluation hyperparameters.}
\label{tab:eval-hparams}
\begin{tabular}{lc}
\toprule
\textbf{Hyperparameter} & \textbf{Value} \\
\midrule
Samples per problem ($n$) & 8 \\
Temperature               & 0.6 \\
Top-$p$                   & 0.95 \\
Top-$k$                   & 20 \\
Max prompt length         & 1536 \\
Max response length       & 3072 \\
\bottomrule
\end{tabular}
\end{table}
\section{Artifacts and licenses}
The datasets, target and teacher models, and sentence encoder used in this work are publicly released research artifacts, used here solely for non-commercial research consistent with their intended use.
Among the datasets, \texttt{MATH}, \texttt{GSM8K}, \texttt{OlympiadBench}, and \texttt{MuSR} are distributed under the MIT License;  and \texttt{FOLIO} under the Creative Commons Attribution-NonCommercial-ShareAlike 4.0 (CC BY-NC-SA 4.0).
Among the models, \texttt{Llama-3.2-3B-Instruct} is used under the Llama 3.2 Community License; \texttt{Qwen-2.5-1.5B-Instruct}, \texttt{Phi-3.5-mini-instruct}, and \texttt{all-MiniLM-L6-v2} under Apache 2.0; and \texttt{gemma-2-2b-it} and \texttt{gemma-3-3b-it} under the Gemma Terms of Use. \texttt{DeepSeek-V4-Flash} is accessed through the OpenRouter API under the provider's terms of service. The datasets consist of competition mathematics, grade-school math, and logical-reasoning problems, and contain no personally identifying information or offensive content.
\definecolor{refbg}{RGB}{255,249,219}
\definecolor{refborder}{RGB}{204,153,0}
\definecolor{outputbg}{RGB}{242,250,244}
\definecolor{outputborder}{RGB}{76,153,102}

\newtcblisting{systemprompt}[1]{
  breakable,
  listing only,
  title={\small\bfseries\sffamily #1},
  colback=refbg,
  colframe=refborder,
  colbacktitle=refborder!80!white,
  coltitle=white,
  arc=3pt,
  boxrule=0.6pt,
  leftrule=3pt,
  listing options={
    basicstyle=\ttfamily\small,
    breaklines=true,
    breakatwhitespace=false,
    columns=fullflexible,
    keepspaces=true,
  },
}

\newtcblisting{userprompt}[1]{
  breakable,
  listing only,
  title={\small\bfseries\sffamily #1},
  colback=outputbg,
  colframe=outputborder,
  colbacktitle=outputborder!80!white,
  coltitle=white,
  arc=3pt,
  boxrule=0.6pt,
  leftrule=3pt,
  listing options={
    basicstyle=\ttfamily\small,
    breaklines=true,
    breakatwhitespace=false,
    columns=fullflexible,
    keepspaces=true,
  },
}
\section{Prompts used in the paper}
\subsection{Prompt for extracting abstractions from traces}
\label{app:prompt-extract-abs}
\begin{systemprompt}{System prompt}
You are extracting reusable mathematical reasoning abstractions from a model trace.

Your job is to read one math problem, the ground-truth final answer, one model-generated trace, and a correctness label telling you whether the trace is correct or incorrect.

Produce 1 or 2 reusable abstractions only. Each abstraction must be:
- a single sentence in "principle"
- generic and reusable across similar problems
- free of problem-specific numbers, variable names, answer values, or story details
- phrased as reasoning advice, not as a restatement of the solution

Allowed abstraction types:
- "strategy": a positive reasoning move that should be reused
- "caution": a mistake pattern or failure mode that should be avoided

Output a JSON array with 1 or 2 objects. Each object must contain exactly these keys:
- "name"
- "type"
- "principle"
- "when_to_apply"
- "domain"

Important rules:
- Do not copy phrases from the trace unless necessary.
- Do not mention the final answer explicitly.
- Do not mention concrete numbers unless they are universal constants like 0, 1, or pi.
- If the trace is correct, prefer strategy abstractions, but you may include one caution if the reasoning contains a risky pattern.
- If the trace is incorrect, prefer caution abstractions, but you may include one strategy if you can clearly infer a better reasoning move from the problem and the ground-truth answer.
- If there is only one strong abstraction, output only one.
\end{systemprompt}

\begin{userprompt}{User prompt for correct problems}
Problem:
{problem_text}

Ground-truth final answer:
{ground_truth}

Model trace:
{trace_text}

Correctness label:
{label}

Task:
Extract 1 or 2 reusable one-sentence abstractions from this example.

Additional guidance:
This trace reached the correct final answer. Focus on the key reasoning move(s) worth reusing in future problems. Prefer strategy abstractions. You may include one caution only if the trace contains a potentially dangerous shortcut that happened to work here.

Return only a JSON array.
\end{userprompt}

\begin{userprompt}{User prompt for incorrect problems}
Problem:
{problem_text}

Ground-truth final answer:
{ground_truth}

Model trace:
{trace_text}

Correctness label:
{label}

Task:
Extract 1 or 2 reusable one-sentence abstractions from this example.

Additional guidance:
This trace did not reach the correct final answer. Focus on the main mistake pattern that should be avoided in future problems. Prefer caution abstractions. You may include one strategy only if a clearly better reasoning direction is recoverable from the problem and the ground-truth answer.

Return only a JSON array.
\end{userprompt}

\subsection{Prompt for rewriting input problems}
\label{app:prompt-rewrite-input}
\begin{systemprompt}{System prompt}
You are a math tutor writing retrieval abstractions. Output exactly one line, one sentence, in this format:
<Type>: <short abstraction>, where <Type> is either Strategy or Caution. For example \"Check extraneous roots\". Do not solve the problem and do not provide a final answer.
\end{systemprompt}

\begin{userprompt}{User prompt}
    Problem: 
    {text}
    
    Abstraction (Strategy or Caution):
\end{userprompt}

\subsection{Prompt for problem solving}
\label{app:prompt-abs-rollout}
\begin{systemprompt}{System prompt for abstraction-augmented problem-solving}
    <reference_notes>
The following notes were retrieved from a library of lessons distilled
from past attempts on similar problems. They MAY OR MAY NOT apply to
the current problem.

How to use them:
- STRATEGIES are candidate approaches - adopt one only if it directly fits.
- CAUTIONS are common errors - before giving the final answer, verify
  you are not making any that apply here.
- If none apply, solve the problem normally.
- Properly cite the retrieved notes in your answer.

## Strategies
- When simplifying an expression, combine like terms by performing the indicated arithmetic operation on their coefficients.
  when: When simplifying linear algebraic expressions involving addition or subtraction of terms with the same variable and exponent.
- Factor quadratic equations when possible to find their roots.
  when: When a quadratic can be factored over the integers.

## Cautions (verify at the end)
- When solving inequalities involving fractions, ensure the denominator is not zero.
  when: When solving rational inequalities or equations with variable denominators.
</reference_notes>

Please reason step by step, and put your final answer within \boxed{}.
\end{systemprompt}

\begin{systemprompt}{Vanilla system prompt}
Please reason step by step, and put your final answer within \boxed{}.
\end{systemprompt}

\begin{userprompt}{User prompt}
    {Problem text}
\end{userprompt}

\section{Example inference traces on MATH-500 with abstractions}
\definecolor{promptbg}{RGB}{245,247,250}
\definecolor{refbg}{RGB}{255,249,219}
\definecolor{refborder}{RGB}{204,153,0}
\definecolor{outputbg}{RGB}{242,250,244}
\definecolor{outputborder}{RGB}{76,153,102}
\definecolor{syslabel}{RGB}{100,100,120}

\newtcolorbox{promptbox}[1]{
  breakable,
  colback=promptbg,
  colframe=gray!50,
  fonttitle=\bfseries\small,
  title=#1,
  left=6pt, right=6pt, top=4pt, bottom=4pt,
  boxrule=0.5pt,
}

\newtcolorbox{refbox}{
  breakable,
  colback=refbg,
  colframe=refborder,
  boxrule=0.8pt,
  leftrule=3pt,
  left=6pt, right=6pt, top=3pt, bottom=3pt,
  fontupper=\small\ttfamily,
}

\newtcolorbox{outputbox}{
  breakable,
  colback=outputbg,
  colframe=outputborder,
  boxrule=0.5pt,
  leftrule=3pt,
  left=6pt, right=6pt, top=3pt, bottom=3pt,
  fontupper=\small,
}

\newcommand{\rolelabel}[1]{%
  {\scriptsize\sffamily\color{syslabel}\textbf{[#1]}}\enspace}

\begin{figure*}[ht]
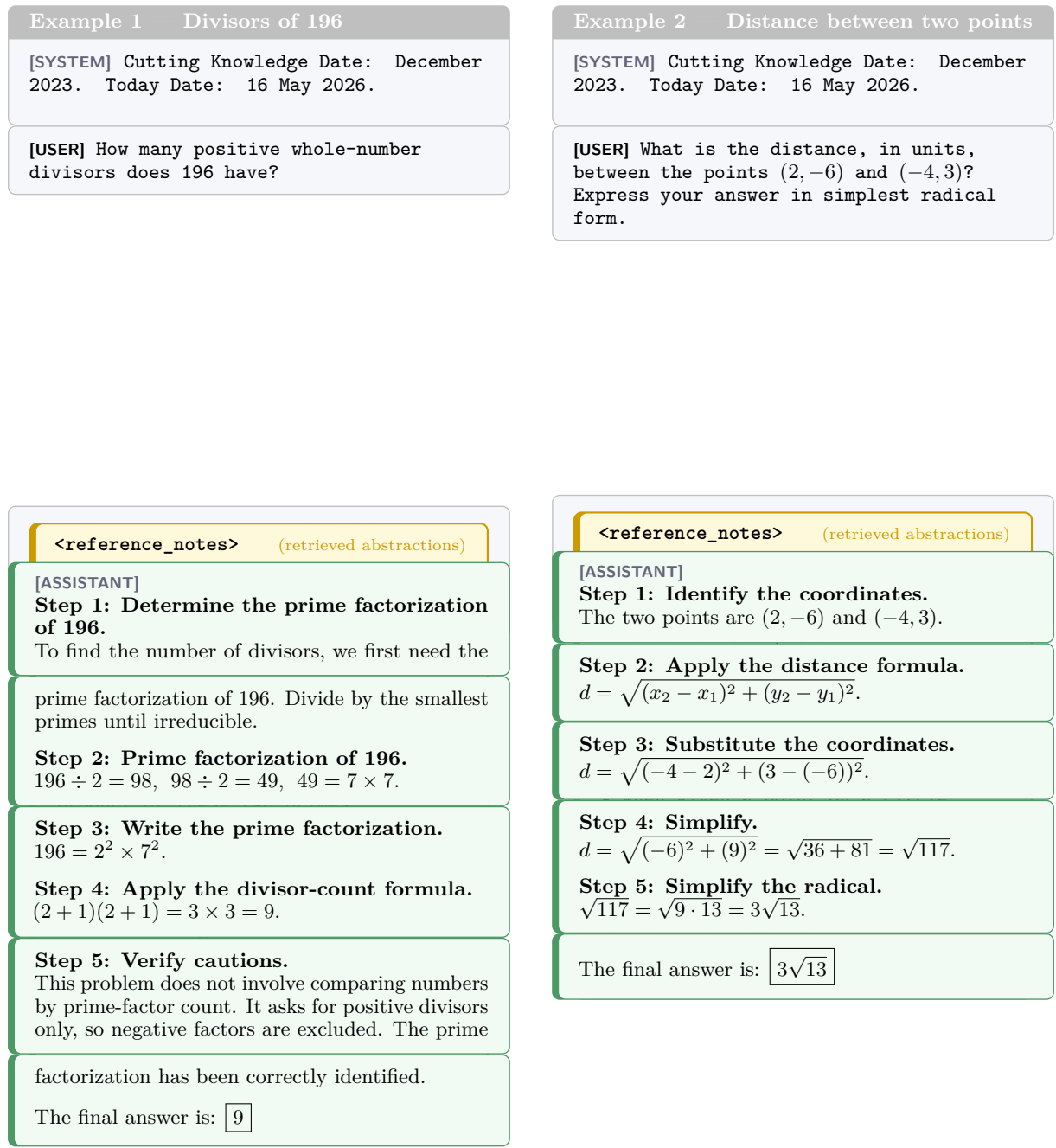

\begin{minipage}[t]{0.48\textwidth}

\begin{promptbox}{Example 1 --- Divisors of 196}
\rolelabel{SYSTEM}\small\ttfamily
Cutting Knowledge Date: December 2023. Today Date: 16 May 2026.

\vspace{6pt}
\rolelabel{USER}\small\ttfamily
How many positive whole-number divisors does 196 have?

\vspace{4pt}
\begin{refbox}
\textbf{<reference\_notes>}\hfill{\rmfamily\scriptsize\color{refborder}(retrieved abstractions)}\\[3pt]
\textbf{Strategies}\\
\textbullet~When determining the number of divisors of a number, use its prime factorization to calculate the number of divisors.\\
\textbullet~A number has exactly three positive divisors if and only if it is the square of a prime number.\\
\textbullet~The next divisor must be a multiple of either 17 or 19, as 323 is a product of these two primes.\\[4pt]
\textbf{Cautions}\\
\textbullet~A number with more prime factors does not necessarily have more divisors.\\
\textbullet~When finding divisors, consider both positive and negative factors.\\
\textbullet~Ensure the prime factorization is correct before applying the divisor formula.\\
\textbf{</reference\_notes>}
\end{refbox}
\end{promptbox}
\begin{outputbox}
\rolelabel{ASSISTANT}\rmfamily\small

\textbf{Step 1: Determine the prime factorization of 196.}\\
To find the number of divisors, we first need the prime factorization of 196. Divide by the smallest primes until irreducible.

\medskip
\textbf{Step 2: Prime factorization of 196.}\\
$196 \div 2 = 98$,\; $98 \div 2 = 49$,\; $49 = 7 \times 7$.

\medskip
\textbf{Step 3: Write the prime factorization.}\\
$196 = 2^2 \times 7^2$.

\medskip
\textbf{Step 4: Apply the divisor-count formula.}\\
$(2+1)(2+1) = 3 \times 3 = 9$.

\medskip
\textbf{Step 5: Verify cautions.}\\
This problem does not involve comparing numbers by prime-factor count. It asks for positive divisors only, so negative factors are excluded. The prime factorization has been correctly identified.

\medskip
The final answer is: $\boxed{9}$
\end{outputbox}

\end{minipage}
\hfill
\begin{minipage}[t]{0.48\textwidth}

\begin{promptbox}{Example 2 --- Distance between two points}
\rolelabel{SYSTEM}\small\ttfamily
Cutting Knowledge Date: December 2023. Today Date: 16 May 2026.

\vspace{6pt}
\rolelabel{USER}\small\ttfamily
What is the distance, in units, between the points $(2, -6)$ and $(-4, 3)$? Express your answer in simplest radical form.

\vspace{4pt}
\begin{refbox}
\textbf{<reference\_notes>}\hfill{\rmfamily\scriptsize\color{refborder}(retrieved abstractions)}\\[3pt]
\textbf{Strategies}\\
\textbullet~Use $d = \sqrt{(x_2-x_1)^2+(y_2-y_1)^2}$ to find the distance between two points in a coordinate plane.\\
\textbullet~When given speed and time, use $\text{distance} = \text{speed} \times \text{time}$.\\
\textbullet~Use the distance formula for two points in 3D space.\\
\textbullet~For equidistant points, set up an equation via the distance formula.\\
\textbullet~When given a point on a circle, use the distance formula to find the radius.\\[4pt]
\textbf{Cautions}\\
\textbullet~In the complex plane, use $\sqrt{(x_2-x_1)^2+(y_2-y_1)^2}$ rather than $|z_2 - z_1|$.\\
\textbf{</reference\_notes>}
\end{refbox}
\end{promptbox}
\begin{outputbox}
\rolelabel{ASSISTANT}\rmfamily\small

\textbf{Step 1: Identify the coordinates.}\\
The two points are $(2, -6)$ and $(-4, 3)$.

\medskip
\textbf{Step 2: Apply the distance formula.}\\
$d = \sqrt{(x_2-x_1)^2+(y_2-y_1)^2}$.

\medskip
\textbf{Step 3: Substitute the coordinates.}\\
$d = \sqrt{(-4-2)^2+(3-(-6))^2}$.

\medskip
\textbf{Step 4: Simplify.}\\
$d = \sqrt{(-6)^2+(9)^2} = \sqrt{36+81} = \sqrt{117}$.

\medskip
\textbf{Step 5: Simplify the radical.}\\
$\sqrt{117} = \sqrt{9 \cdot 13} = 3\sqrt{13}$.

\medskip
The final answer is: $\boxed{3\sqrt{13}}$
\end{outputbox}

\end{minipage}

\caption{Two inference examples in the MATH-500 test dataset. The \colorbox{refbg}{yellow-shaded block} in each prompt
is the retrieved abstraction injected at inference time. Note that the
user question is shown first for readability; the abstraction block follows.}
\label{fig:examples}
\end{figure*}

Figure~\ref{fig:examples} shows two complete inference traces on \texttt{MATH-500}, with the retrieved abstraction block highlighted in yellow.

\end{document}